\documentclass[
]{ceurart}

\usepackage{listings}
\usepackage{xcolor}
\usepackage{soul}

\usepackage{amsmath}
\usepackage{bm}
\usepackage{amssymb}
\usepackage{dsfont}
\usepackage{siunitx}
\graphicspath{{./}}
\usepackage{caption}
\usepackage{subcaption}
\usepackage{multirow}
\usepackage{makecell}
\usepackage{arydshln}

\usepackage{bbding}			
\usepackage{makecell}
\usepackage{arydshln}		
\usepackage{makecell}		

\newcommand{\NA}{---}
\DeclareMathOperator*{\argmax}{arg\,max}
\newcommand{\gray}[1]{\textcolor{gray}{#1}}

\begin{document}

\copyrightyear{2022}
\copyrightclause{Copyright for this paper by its authors.
  Use permitted under Creative Commons License Attribution 4.0
  International (CC BY 4.0).}

\conference{EvalLAC'25: 2nd Workshop on Automatic Evaluation of Learning and Assessment Content, July 26, 2025, Palermo, Italy}

\title{Ordinality in Discrete-level Question Difficulty Estimation: Introducing Balanced DRPS and OrderedLogitNN}

\author[1,2]{Arthur Thuy}[%
orcid=0000-0001-9107-5646,
email=arthur.thuy@ugent.be,
]
\cormark[1]
\address[1]{Ghent University, Tweekerkenstraat 2, 9000 Ghent, Belgium}
\address[2]{CVAMO Core Lab Flanders Make, Tweekerkenstraat 2, 9000 Ghent, Belgium}
\address[3]{Dedalus Healthcare, Roderveldlaan 2, 2600 Antwerp, Belgium}

\author[3]{Ekaterina Loginova}[%
orcid=0000-0002-0952-9213,
email=ekaterina.d.loginova@gmail.com,
]

\author[1,2]{Dries F.~Benoit}[%
orcid=0000-0002-9421-8566,
email=dries.benoit@ugent.be,
]

\cortext[1]{Corresponding author.}




\begin{abstract}
	Recent years have seen growing interest in Question Difficulty Estimation (QDE) using natural language processing techniques. 
	Question difficulty is often represented using discrete levels, framing the task as ordinal regression due to the inherent ordering from easiest to hardest.
	However, the literature has neglected the ordinal nature of the task, relying on classification or discretized regression models, with specialized ordinal regression methods remaining unexplored. 
	Furthermore, evaluation metrics are tightly coupled to the modeling paradigm, hindering cross-study comparability.
	While some metrics fail to account for the ordinal structure of difficulty levels, none adequately address class imbalance, resulting in biased performance assessments.
	This study addresses these limitations by benchmarking three types of model outputs---discretized regression, classification, and ordinal regression---using the balanced Discrete Ranked Probability Score (DRPS), a novel metric that jointly captures ordinality and class imbalance.	
	In addition to using popular ordinal regression methods, we propose OrderedLogitNN, extending the ordered logit model from econometrics to neural networks.
	We fine-tune BERT on the RACE\texttt{++} and ARC datasets and find that OrderedLogitNN performs considerably better on complex tasks.
	The balanced DRPS offers a robust and fair evaluation metric for discrete-level QDE, providing a principled foundation for future research.
\end{abstract}


\maketitle


\section{Introduction}
\label{sec:intro}

Question Difficulty Estimation (QDE), also known as question calibration, aims to predict a question's difficulty directly from its textual content and its answer options. 
This task plays a central role in personalized learning tools such as computerized adaptive testing \cite{van2000computerized} and dynamic online learning platforms, which aim to present questions aligned with a learner’s proficiency level. 
Selecting questions that are either too easy or excessively difficult can reduce student motivation and hinder learning outcomes \cite{wang2014regularized}. 
Reliable estimation of question difficulty is therefore essential.

Traditionally, QDE has relied on manual calibration \cite{attali2014estimating} or pretesting \cite{lane2016handbook}, both of which are time-consuming and costly. 
To address these limitations, research has explored the use of natural language processing (NLP) techniques. 
These approaches train machine learning models to infer difficulty from the question text, allowing for rapid and scalable calibration of new questions without the need for manual intervention.

Difficulty levels in QDE are represented either as continuous scores or as discrete categories, with discrete levels being attractive for their ease of use.
While continuous difficulty estimation is generally framed as a regression task, discrete-level QDE is essentially an ordinal regression problem, given the inherent ordering of difficulty levels from easiest to hardest.

However, the discrete-level QDE literature has neglected the ordinal nature of the task. 
Instead, existing work relies exclusively on classification and discretized regression methods, both of which are oversimplifications of the problem structure. 
Classification models disregard ordinal relationships altogether, and while discretized regression methods preserve ordinality, they implicitly assume equal spacing between levels---an assumption often violated in real-world data.
As such, specialized ordinal regression techniques remain unexplored in this context.
Moreover, no prior studies have systematically compared these competing approaches.
Compounding the issue, studies typically only report performance metrics aligning with their chosen modeling paradigm, making cross-study comparisons difficult.
The metrics also fail to account for class imbalances, a prevalent issue in this setting.
As a result, there is no consensus on the most effective evaluation metric or modeling approach for discrete-level QDE.

This study addresses the literature gaps in discrete-level QDE by proposing the balanced Discrete Ranked Probability Score (DRPS), a novel evaluation metric that jointly captures ordinality and class imbalance.
It also provides a direct way to compare deterministic predictions to probabilistic ones, which are especially valuable for downstream decision-making.
We benchmark three types of model outputs---discretized regression, classification, and ordinal regression---using the balanced DRPS.
Moreover, we propose a novel ordinal regression model OrderedLogitNN, extending the ordered logit model from econometrics to neural networks (NNs).
Our work is the first to (i) introduce the balanced DRPS metric, (ii) compare classification and discretized regression models for QDE, and (iii) investigate specialized ordinal regression techniques, including the novel OrderedLogitNN.
We conduct experiments by fine-tuning the Transformer model BERT on the RACE\texttt{++} and ARC datasets.

The remainder of the paper is structured as follows. 
Section \ref{sec:related_work} reviews related work.
Section \ref{sec:drps} introduces the balanced DRPS metric. 
Section \ref{sec:orderedlogitnn} discusses the novel OrderedLogitNN model and section \ref{sec:model_output} outlines existing methods for ordinal tasks. 
Section \ref{sec:experiments} describes our experimental setup, and Section \ref{sec:results_discussion} presents the results and discussion. 
We conclude in Section \ref{sec:conclusion}.
The source code is available on GitHub.\footnote{\url{https://github.com/arthur-thuy/qde-ordinality}}


\section{Related Work}
\label{sec:related_work}

Building on the survey by \cite{benedetto2023survey}, we further investigate studies in QDE that utilize datasets with discrete difficulty levels.
Question difficulty is defined using one of three main approaches: (i) Classical Test Theory (CTT) \cite{hambleton1993comparison}, (ii) Item Response Theory (IRT) \cite{hambleton1991fundamentals}, and (iii) manual calibration.
In the case of manual calibration with expert annotators, difficulty is almost exclusively assigned in discrete levels due to its ease of use.
While difficulty scores derived from CTT and IRT are continuous by nature, they are often discretized in practical applications to facilitate interpretation.
Table~\ref{tab:related_work} provides an overview of related work in discrete-level QDE.

\begin{table}[t!]
	\caption{Related work on discrete-leveled QDE}\label{tab:related_work}
	\begin{tabular}{ccccccc}  
		\toprule%
		\multirow{2}{*}{\raisebox{-\heavyrulewidth}{Paper}} & 
		\multirow{2}{*}{\raisebox{-\heavyrulewidth}{Year}} & \multirow{2}{*}{\raisebox{-\heavyrulewidth}{Difficulty}} & \multicolumn{3}{c}{Model output format} &
		\multirow{2}{*}{\raisebox{-\heavyrulewidth}{Metric}}\\
		\cmidrule{4-6}
		& & & Regression & Classification & Ordinal & \\
		\midrule
		\cite{hsu2018automated} & 2018 & IRT & \NA & \Checkmark & \NA & (Adjacent) accuracy\\
		\cite{yang2018feature} & 2018 & Manual & \NA & \Checkmark & \NA & Accuracy\\
		\cite{fang2019exercise} & 2019 & Manual & \NA & \Checkmark & \NA & Accuracy\\
		\cite{lin2019automated} & 2019 & Manual, IRT & \Checkmark & \NA & \NA & (Adjacent) accuracy\\
		\cite{zhou2020multi} & 2020 & CTT & \NA & \Checkmark & \NA & Accuracy, $F_1$-score\\
		\cite{loginova2021towards} & 2021 & Manual & \NA & \Checkmark & \NA & Accuracy\\ 
		\cite{benedetto2023quantitative} & 2023 & Manual & \Checkmark & \NA & \NA & RMSE, $R^2$, Spearman's $\rho$\\
		\cite{thuy2024active} & 2024 & Manual & \Checkmark & \NA & \NA & RMSE\\
		\midrule
		\textbf{Current} & 2025 & Manual & \Checkmark & \Checkmark & \Checkmark & Balanced DRPS\\
		\bottomrule
	\end{tabular}
\end{table}

\subsection{Output types}

Early work on discrete-level QDE employed classification models, such as support vector machines and Bayesian NNs \cite{hsu2018automated, yang2018feature, fang2019exercise}. 
Subsequently, \cite{lin2019automated} proposed a discretized regression approach using an LSTM-based NN, where difficulty was predicted as a continuous value between 0 and 1 and mapped to discrete intervals (e.g., $[0.0;0.2)$, $[0.2;0.4)$, etc.).

\cite{zhou2020multi} introduced a multi-task BERT-based model that leverages shared representations across datasets, using a classification head to predict difficulty levels. 
To reduce the reliance on large labeled datasets in supervised methods, \cite{loginova2021towards} proposed an unsupervised QDE method. 
They leverage the uncertainty in pre-trained question-answering models as a proxy for human-perceived difficulty, computed as the variance over the predictions from an ensemble of classification models.

\cite{benedetto2023quantitative} conducted a benchmarking study comparing traditional machine learning methods and end-to-end NNs across datasets with both discrete and continuous difficulty labels. 
Their results show that fine-tuned Transformer-based models such as BERT and DistilBERT consistently outperform classical methods. 
However, these models exclusively employed a discretized regression approach to address the ordinal nature of the labels.

More recently, \cite{thuy2024active} investigated active learning for QDE, demonstrating that comparable performance to fully supervised models can be achieved by labeling only a small fraction of the training data. 
Yet again, only a discretized regression modeling strategy was considered. 
As summarized in Table~\ref{tab:related_work}, prior work has not directly compared discretized regression and classification approaches, and specialized ordinal regression methods remain entirely unexplored.

\subsection{Metrics}
\label{subsec:metrics}

Accuracy is the most widely used evaluation metric for discrete-level QDE, particularly in studies employing classification models, as it aligns with the cross-entropy loss typically used during training. 
\cite{lin2019automated} is the only study using a discretized regression approach that also reports accuracy. 
However, accuracy fails to account for the ordinal structure of difficulty levels: all misclassifications are treated equally, regardless of their distance from the true label.
Even if a prediction is incorrect, it should still be as close as possible to the true difficulty level.
Additionally, accuracy is a threshold-based metric, relying solely on the final predicted label rather than the full output distribution---an issue that also applies to metrics such as the $F_1$-score.
To partially address this, \cite{hsu2018automated} and \cite{lin2019automated} report adjacent accuracy, defined as the proportion of predictions within $k$ levels of the true label. 
However, the choice of $k$ is often arbitrary and dataset-dependent, complicating comparisons across studies.

The most recent works, which adopt discretized regression approaches, report RMSE as their primary evaluation metric, treating difficulty levels as integers. 
While RMSE reflects the ordinal structure to some extent, it assumes uniform spacing between levels---a condition rarely met in real-world data.
For example, in primary school, there is a non-linear increase in difficulty over the years \cite{coe2008relative}.
This limitation also applies to metrics such as $R^2$ and Spearman’s rank correlation. 
Furthermore, since RMSE is used as the loss function in discretized regression models, these models benefit from being evaluated on the same objective they were optimized for, giving them an unfair advantage and introducing a potential bias in comparative evaluations.

In addition to the ordinality aspect, these commonly used metrics fail to account for class imbalance, which is a prevalent issue in discrete-level QDE. 
In many datasets, mid-range difficulty levels tend to dominate, while questions at the extremes---those that are very easy or very difficult---are underrepresented \cite{liang2019new, clark2018think}. 
Standard metrics, which compute aggregate scores across all samples, inherently place greater weight on the majority classes. 
As a result, model performance on minority classes is often underrepresented, leading to inflated metrics that do not accurately reflect model effectiveness across the full difficulty spectrum. 
This is particularly problematic in educational contexts, where balanced performance across all difficulty levels is essential to ensure adequate personalized learning experiences.

As a result, the literature ignores the ordinal aspect in the modeling approaches and employs suboptimal evaluation metrics for discrete-level QDE. 
There is a clear need for an evaluation metric that simultaneously accounts for ordinality and class imbalance, thereby enabling fair comparison across different modeling strategies.

\section{Balanced Discrete Ranked Probability Score}
\label{sec:drps}

The Continuous Ranked Probability Score (CRPS) is the most widely adopted scoring rule for evaluating probabilistic forecasts of real-valued variables, such as in precipitation forecasting \cite{gneiting2007strictly}.
It is defined as the integral of the squared difference (i.e., Brier score) between the cumulative distribution function (CDF) of a probabilistic forecast $F$ and the CDF of the observed outcome, at all real-valued thresholds.
The observed outcome is represented as a degenerate distribution, as its CDF is a step function.
Formally, given a dataset $D = \{\mathbf{x}_i, y_i\}^{N}_{i=1}$, the CRPS is computed as:
\begin{equation*}
	\text{CRPS}(F, y) = \frac{1}{N}\sum_{i=1}^{N} \int_{-\infty}^{\infty} (F(\hat{y_i}) - \mathds{1}\{\hat{y}_i \geq y_i\})^2 \,d\hat{y}_i\,,
\end{equation*}
where $F(\hat{y_i})$ denotes the CDF of the forecast and $\mathds{1}\{\cdot\}$ is the step function.

The CRPS is distance-sensitive, meaning it rewards forecasts that assign higher probability mass to values near the true outcome. 
Specifically, when the forecast distribution concentrates probability density around the true value, the squared error between the forecast CDF and the observed step function is smaller across the integration range, resulting in a lower (better) score. 
In other words, even if the predicted value does not exactly coincide with the true outcome, placing substantial probability on neighboring values results in a better score than a prediction that is entirely off-target.

This property makes the CRPS particularly well-suited for ordinal prediction tasks. 
In such cases, the DRPS serves as a natural extension of the CRPS for discrete outcomes across $K$ ordered categories. 
The DRPS has only been applied in meteorology \cite{weigel2007discrete} and has received little attention in the general field of ordinal regression.
It is defined as:
\begin{equation*}
	\text{DRPS}(F, y) = \frac{1}{N}\sum_{i=1}^{N} \sum_{k=1}^{K-1} (F_k(\hat{y}_i) - \mathds{1}\{k \geq y_i\})^2\,,
\end{equation*}
where $F_k(\hat{y}_i)$ denotes the predicted cumulative probability up to class $k$.
The step function $\mathds{1}\{\cdot\}$ moves from 0.0 to 1.0 at the position of the ground truth label.

\begin{figure}[t!]
	\centering
	\begin{subfigure}[b]{0.50\textwidth}
		\centering
		\includegraphics[width=1.0\linewidth]{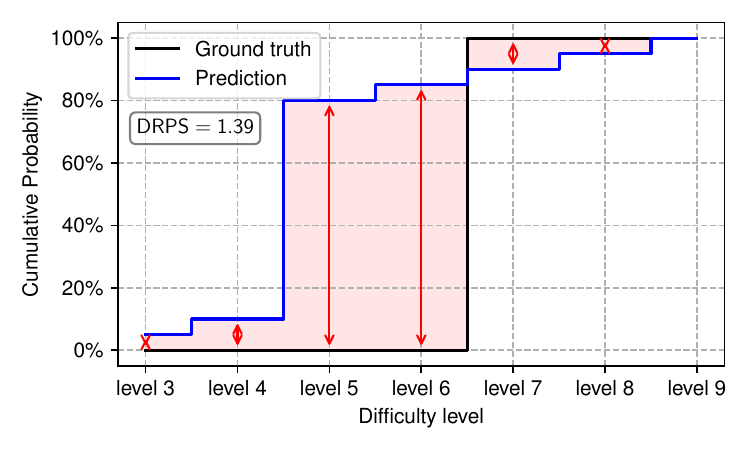}
		\caption{Most density on ``level 5'', confident}
		\label{fig:drps_example_probabilities_certain} 
	\end{subfigure}%
	\begin{subfigure}[b]{0.50\textwidth}
		\centering
		\includegraphics[width=1.0\linewidth]{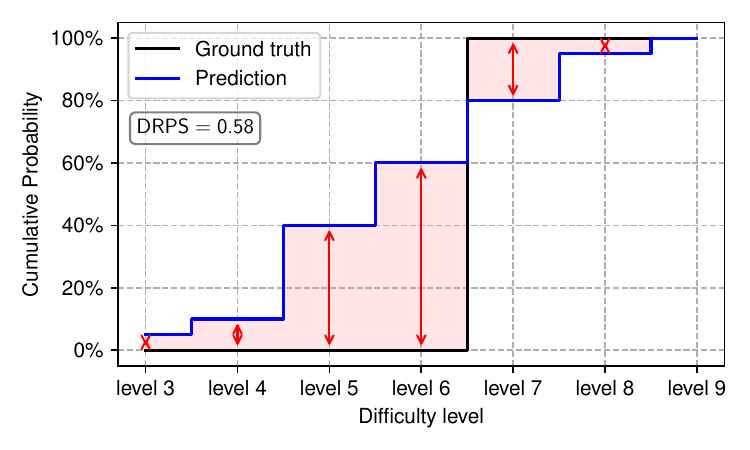}
		\caption{Most density on ``level 5'', uncertain}
		\label{fig:drps_example_probabilities_uncertain}
	\end{subfigure}
	
	\caption{Example DRPS calculation with predicted probabilities. The ground truth for this observation is level 7 and both predictions have the most density on the incorrect level 5. (a) assigns high probability to level 5 while (b) is more uncertain and has more density on the neighboring levels, resulting in a better score.}
	\label{fig:drps_example_probabilities}
\end{figure}

\begin{figure}[t!]
	\centering
	\begin{subfigure}[b]{0.50\textwidth}
		\centering
		\includegraphics[width=1.0\linewidth]{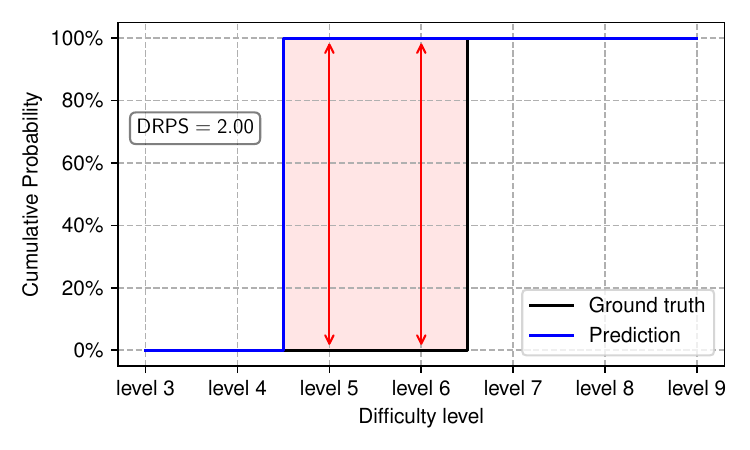}
		\caption{Prediction ``level 5''}
		\label{fig:drps_example_label_minus2} 
	\end{subfigure}%
	\begin{subfigure}[b]{0.50\textwidth}
		\centering
		\includegraphics[width=1.0\linewidth]{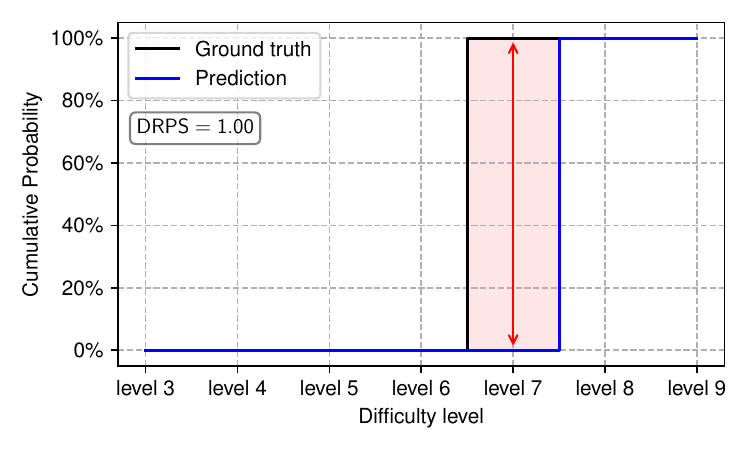}
		\caption{Prediction ``level 8''}
		\label{fig:drps_example_label_plus1}
	\end{subfigure}
	
	\caption{Example DRPS calculation with a predicted label. The ground truth for this observation is level 7; (a) predicts level 5 while (b) predicts level 8.}
	\label{fig:drps_example_label}
\end{figure}

Unlike other evaluation metrics in related work, the DRPS operates on full probability distributions rather than point estimates, enabling a more nuanced assessment of ordinal predictions.
Such a probability distribution over the levels is available for all classification and ordinal regression models, but not for the discretized regression model as it only outputs a predicted difficulty level.
In the case of deterministic predictions, the output is treated as a degenerate distribution---analogous to the representation of the observed outcome.
Figures \ref{fig:drps_example_probabilities} and \ref{fig:drps_example_label} illustrate how the DRPS is computed for single observations with probabilistic and deterministic model outputs.

Crucially, the DRPS respects the ordinal structure of the prediction task without assuming equal inter-class distances, a notable limitation of metrics such as RMSE. 
Additionally, when applied to deterministic predictions, the DRPS reduces to the mean absolute error, providing a direct way to compare deterministic and probabilistic predictions within a unified evaluation metric.

In this work, we introduce the balanced DRPS to address the class imbalance in discrete-level QDE datasets, where extreme difficulty levels are typically underrepresented compared to mid-range levels \cite{liang2019new,clark2018think}.
The popular metrics (accuracy and RMSE) and standard DRPS compute unweighted averages, which overemphasize performance on majority classes and can produce misleadingly high scores on imbalanced data.
However, for educational practitioners, robust performance across the full spectrum of difficulty levels---including the rarest---is essential.
To ensure fair evaluation, the balanced DRPS weights each observation inversely proportional to the prevalence of its true class, $w_i = \frac{1}{\sum_{j=1}^{N} \mathds{1}\{y_j = y_i\}}$, thereby giving equal importance to all difficulty levels.
Formally:
\begin{equation*}
	\text{Balanced DRPS}(F, y) = \frac{1}{N}\sum_{i=1}^{N} \sum_{k=1}^{K-1} w_i \left(F_k(\hat{y}_{i}) - \mathds{1}\{k \geq y_i\}\right)^2.
\end{equation*}
Thus for balanced datasets, the balanced DRPS is equivalent to the standard DRPS.

In conclusion, the balanced DRPS offers a robust and fair evaluation metric for discrete-level QDE by accounting for both ordinal structure and class imbalance. 
It supports both deterministic and probabilistic predictions and remains neutral to training objectives, making it well-suited for benchmarking across diverse modeling approaches.

\section{Ordered Logit for NNs}
\label{sec:orderedlogitnn}

OrderedLogitNN extends the classical ordered logit model to NNs, effectively bridging the gap between econometrics and deep learning. 
At its core, it is a latent variable model, where each observation is associated with an unobserved continuous utility value $y_i^*$ \cite{greene2010modeling} modeled as: $y_i^* = \mathbf{x}_i \mathbf{\beta}  + \epsilon_i$.
The observed ordinal outcome $y_i$ is derived from $y_i^*$ through a censoring mechanism, whereby the continuous latent variable is mapped to one of $K$ discrete categories based on a sequence of $K+1$ increasing threshold values $\{\mu_{-1}, \mu_0, \mu_1, \dots, \mu_{K-1} \}$.

To identify the model parameters, several normalizations are required.
First, the thresholds must be increasing $\mu_k > \mu_{k-1}$ to ensure valid (i.e., positive) probabilities.
Second, the endpoints of the support are fixed as $\mu_{-1} = -\infty$ and $\mu_{K-1} = +\infty$, covering the entire real line.
Third, the error term  $\epsilon_i$ is assumed to follow a standardized logistic distribution (mean zero, variance $\frac{\pi^2}{3}$). 
The logistic distribution is preferred over the Gaussian (i.e., probit) for computational convenience, as the derivative has a closed form solution and is readily available as the sigmoid function.
Finally, since $\mathbf{x}_i$ includes a bias term, the threshold $\mu_0 = 0$.

\begin{figure}[t!]
	\centering
	\includegraphics[width=0.9\textwidth]{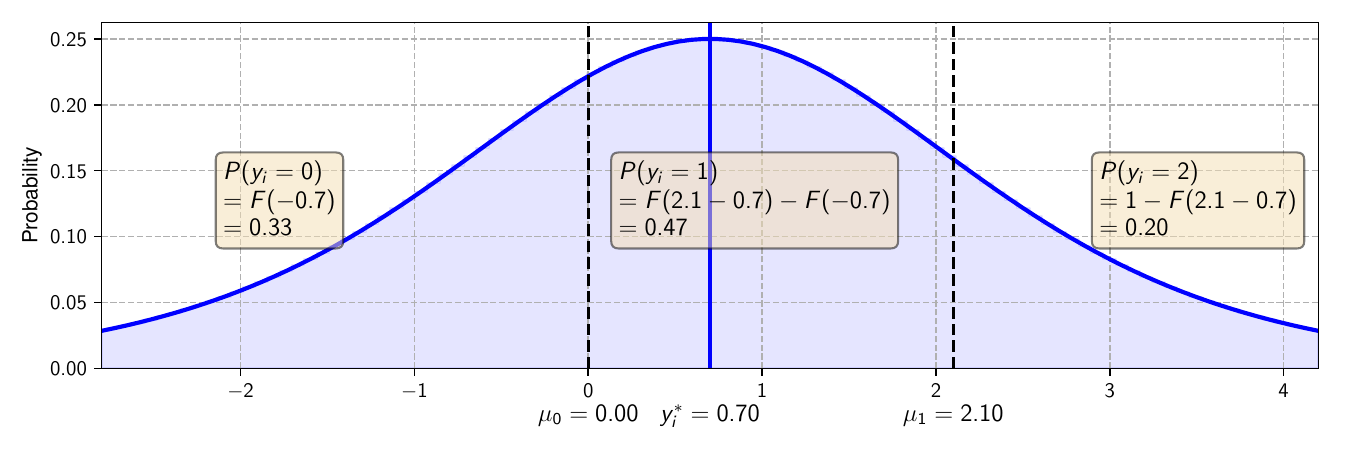}
	\caption{Ordered logit example for three ordinal categories.}
	\label{fig:ordered_logit}
\end{figure}

The model defines the class probabilities as:
\begin{equation*}
	P(y_i = k \mid \mathbf{x}_i) = F(\mu_k - \mathbf{x}_i \mathbf{\beta}) - F(\mu_{k-1} - \mathbf{x}_i \mathbf{\beta})\,,
\end{equation*}
with $F$ the cdf of the logistic distribution.
An example for $K=3$ is shown in Figure \ref{fig:ordered_logit}.

The NN is trained by minimizing the negative log-likelihood (NLL):
\begin{equation*}
	\text{NLL} = \sum_{i=1}^{N} \sum_{j=0}^{K-1} m_{ij} \log \left[ F(\mu_k - \mathbf{x}_i \mathbf{\beta}) - F(\mu_{k-1} - \mathbf{x}_i \mathbf{\beta}) \right]\,,
\end{equation*}
where $m_{ij} = 1$ if $y_i = k$ and 0 otherwise.
The thresholds are reparameterized to ensure monotonically increasing values: $\mu_k = \mu_{k-1} + \exp(\delta_k) = \sum_{m=1}^{k} \exp(\delta_m)$.

The $\delta_k$ parameters are initialized such that the ordinal levels have equal probability mass under the logistic distribution, with the first threshold set to zero. 
The bias term is initialized to lie at the center of this distribution, while all other weights follow standard initialization practices in PyTorch. 
To facilitate convergence, the learning rates for the $\delta_k$ values and the bias term are scaled to be 100 times larger than those of the remaining network parameters.

Importantly, OrderedLogitNN is architecture-agnostic and can be integrated into any NN. 
Additionally, it makes no assumptions about the distances between ordinal levels, allowing it to flexibly model a wide range of ordered regression problems.


\section{Existing approaches for ordinal regression with NNs}
\label{sec:model_output}

This section describes existing approaches to handle ordinal regression using NNs.
Depending on the specific approach, the amount of ordinal information used and the underlying assumptions vary.
This study uses three existing specialized ordinal regression methods that follow the extended binary classification framework, most widely used in the ordinal regression literature \cite{shi2023deep}.
Note that the methods discussed below are not tied to any specific architecture and can be utilized with any NN.

Let $D = \{\mathbf{x}_i, y_i\}^{N}_{i=1}$ be the training dataset consisting of $N$ training examples. Here, $\mathbf{x}_i \in \mathcal{X}$ denotes the $i$\textsuperscript{th} training example and $y_i$ the corresponding rank, where $y_i \in \mathcal{Y} = \{r_1, r_2, \dots, r_K\}$ with ordered rank $r_K \succ r_{K-1} \succ \dots \succ r_1$.
The objective is to find a model that maps $\mathcal{X} \rightarrow \mathcal{Y}$.
For example, the ARC dataset has $K=7$ difficulty levels with an output space $\mathcal{Y} = \{\textrm{``grade 3''}, \textrm{``grade 4''},\dots , \textrm{``grade 9''}\}$.

\subsection{Discretized regression}

In the regression approach, also referred to as discretized regression, the $K$ rank indices are treated as numerical values to utilize the ordinal information (see Table \ref{tab:related_work} for references).
The model $f$ minimizes the mean squared error loss and predicts a real-valued quantity $f(\mathbf{x}_i) \in \mathds{R}$ representing a continuous rank estimate, which is then converted to the closest rank index.
For example, a regression estimate of 2.7 is converted to index 3 while estimate 5.2 is converted to index 5.

Using a discretized regression approach in an ordinal QDE problem assumes that the inter-level distances are equal.
However, this condition is only rarely satisfied in practice \cite{coe2008relative}.
On the ARC dataset with levels ``grade 3'' to ``grade 9'', for example, such an approach assumes that the jump in difficulty among all grades is identical.

\subsection{Classification}

In the classification approach, the model's output space is a set of $K$ unordered labels, one for each rank (see Table \ref{tab:related_work} for references).
The model is trained to minimize the cross-entropy loss and the predicted rank label is the class with the highest predicted probability.
As such, the predicted rank label is $\hat{y}_i = \argmax_{y_i \in \mathcal{Y}} p(y_i \mid \mathbf{x}_i)$.

This approach essentially assumes that the difficulty levels are completely independent, hence discarding the available ordinal information.
For example, for a question in the ARC dataset with true level ``grade 3'', predicting levels ``grade 4'' and ``grade 5'' incurs the same loss even though the difference between ``grade 3'' and ``grade 5'' is larger than the that between level ``grade 3'' and ``grade 4''.

\subsection{Ordinal: OR-NN}

A popular general machine learning approach to ordinal regression is to cast it as an extended binary classification problem \cite{li2006ordinal}, leveraging the relative order among the labels.
That is, the ordinal regression task with $K$ ranks is represented as a series of $K-1$ simpler binary classification sub-problems.
For each rank index $k \in {1, 2, \dots, K - 1}$, a binary classifier is trained according to whether the rank of a sample is larger than $k$.
As such, all $K-1$ tasks share the same intermediate layers but are assigned distinct weight parameters in the output layer.
In summary, this framework relies on three steps: (i) extending rank labels to binary vectors, (ii) training binary classifiers on the extended labels, and (iii) computing the predicted rank label from the binary classifiers.

In 2016, the authors of \cite{niu2016ordinal} adapted this framework to train NNs for ordinal regression; we refer to this method as OR-NN.
More formally, a rank label $y_i$ is first extended into $K-1$ binary vectors $y_i^{(1)}, \dots, y_i^{(K-1)}$ such that the $y_i^{(k)} \in \{0,1\}$ indicates whether $y_i$ exceeds rank $r_k$, for instance, $y_i^{(k)} = \mathds{1}\{y_i > r_k\}$.
Using the extended binary labels, a single NN is trained with $K-1$ binary classifiers in the output layer to minimize the cross-entropy loss.
Based on the binary task predictions, the predicted rank label is $\hat{y}_i  = r_{q_i}$. The rank index $q_i$ is given by
\begin{equation*}
q_i = 1 + \sum_{k=1}^{K-1} \mathds{1}\bigl\{P(y_i > r_k) > 0.5\bigr\}\,,
\end{equation*}
where $P(y_i > r_k) \in [0,1]$ is the predicted probability of the $k$th binary classifiers in the output layer.

\begin{figure}[t!]
\centering
\begin{subfigure}[b]{0.50\textwidth}
	\centering
	\includegraphics[width=1.0\linewidth]{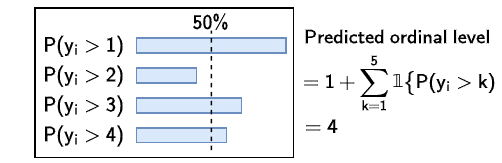}
	\caption{Rank inconsistent}
	\label{fig:rank_inconsistent} 
\end{subfigure}%
\begin{subfigure}[b]{0.50\textwidth}
	\centering
	\includegraphics[width=1.0\linewidth]{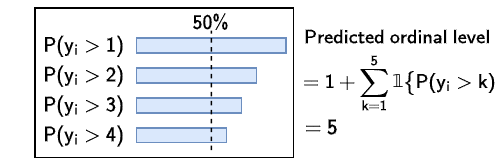}
	\caption{Rank consistent}
	\label{fig:rank_consistent} 
\end{subfigure}

\caption{Example of a rank inconsistent and rank consistent prediction on a QDE task with $K=5$ levels. Figure adapted from \cite{shi2023deep}.}
\label{fig:rank_inconsistent_consistent}
\end{figure}

However, the authors pointed out that OR-NN can suffer from rank inconsistencies among the binary tasks such that the predictions for individual binary tasks may disagree.
For example, on the RACE\texttt{++} dataset, it would be contradictory if the first binary task predicts that the difficulty is not higher than middle school level while the second binary task predicts it to be more difficult than high school level.
This inconsistency could lead to suboptimal results when combining the $K-1$ predictions to obtain the estimated difficulty level.
Figure \ref{fig:rank_inconsistent_consistent} provides an example of a rank consistent and inconsistent prediction.
In response, two methods have been proposed that overcome this drawback of rank inconsistency: CORAL and CORN.

\subsection{Ordinal: CORAL and CORN}

CORAL \cite{cao2020rank} achieves rank consistency by imposing a weight-sharing constraint in the last layer.
Instead of learning distinct weights between each unit in the penultimate layer and each output unit, CORAL enforces that the $K-1$ binary tasks share the same weight parameters to the units in the penultimate layer.
In addition, CORAL learns independent bias terms for each output unit as opposed to a single bias term for the output layer.

CORAL uses a cross-entropy loss over the $K-1$ binary classifiers and the authors show theoretically that by minimizing this loss function, the learned bias terms of the output layer are non-increasing such that $b_1 \geq b_2 \geq \dots \geq b_{K-1}$.
Consequently, the predicted probabilities of the $K-1$ tasks are decreasing which ensures that the output reflects the ordinal information and is rank consistent.
All other steps are identical to the extended binary classification framework.

However, while CORAL outperforms the OR-NN method in age prediction \cite{cao2020rank}, the weight-sharing constraint may restrict the expressiveness and capacity of the NN.

More recently, the authors of \cite{shi2023deep} proposed CORN, which guarantees rank consistency without restricting the NN’s expressiveness and capacity.
CORN achieves rank consistency by a novel training scheme which uses conditional training sets in order to obtain the unconditional rank probabilities.

More formally, CORN constructs conditional training subsets such that the output of the $k$th binary task $f_k(\mathbf{x}_i)$ represents the conditional probability $f_k(\mathbf{x}_i) = P\bigl(y_i > r_k \mid y_i > r_{k-1}\bigr)$.
For $k \geq 2$, the conditional subsets consist of observations where $y_i > r_{k-1}$.
When $k=1$, $f_1(\mathbf{x}_i)$ represents the initial unconditional probability $P\bigl(y_i > r_1\bigr)$ based on the complete dataset.
The transformed unconditional probabilities can then be computed by applying the chain rule for probabilities: $P\bigl(y_i > r_k\bigr) = \prod_{j=1}^{k} f_j(\mathbf{x}_i)$.

Since $\forall j, 0 \leq f_j(\mathbf{x}_i) \leq 1$, we have $P\bigl(y_i > r_1\bigr) \geq P\bigl(y_i > r_2\bigr) \geq \dots \geq P\bigl(y_i > r_{K-1}\bigr)$,
which guarantees rank consistency among the $K-1$ binary tasks.
During model training, CORN minimizes the cross-entropy loss over the binary tasks.
All other steps are identical to the extended binary classification framework.


\section{Experiments}
\label{sec:experiments}

\subsection{Data}

We evaluate our models on two multiple-choice question (MCQ) datasets: RACE\texttt{++} and ARC.
These datasets vary in domain and granularity of difficulty levels.

RACE\texttt{++} \cite{liang2019new,lai2017race} is a dataset of reading comprehension MCQs.
Questions are labeled with one of three difficulty levels---1 (middle school), 2 (high school), and 3 (university level)---which we treat as ground truth labels for QDE. 
The label distribution is imbalanced: 25\% of the questions are labeled as level 1, 62\% as level 2, and 13\% as level 3. 
The dataset is partitioned into training, validation, and test sets containing \num{100568}; \num{5599}; and \num{5642} questions, respectively.

ARC \cite{clark2018think} is a dataset of science MCQs across grades 3 through 9. 
Question difficulty is indicated by the target grade level (i.e., 7 levels), which we use as ground truth. 
The training, validation, and test splits contain \num{3358}, \num{862}, and \num{3530} questions, respectively.
The distribution is highly imbalanced: level 8 appears approximately \num{1400} times, level 5 about 700 times, and level 6 only 100 times. 
To decrease this imbalance, we follow \cite{benedetto2023quantitative} and downsample the two most frequent levels to 500 examples each, resulting in a partially balanced training set of \num{2293} questions.

\subsection{Model Architecture}

We focus on end-to-end Transformer-based NNs, as they have been shown to outperform traditional NLP approaches that rely on separate feature engineering and modeling stages \cite{benedetto2023quantitative}.
We fine-tune the Transformer BERT (``bert-base-uncased'') on the task of QDE, stacking an output layer on top of the pre-trained language model.
During fine-tuning, both the weights of the output head and the pre-trained model are updated.
We follow the input encoding of \cite{benedetto2023quantitative} and concatenate the question and the text of all the possible answer choices in a single sentence, divided by separator tokens.

Additionally, we investigate the performance of two baselines which serve as a lower bound on performance: (i) Random and (ii) Majority. 
The Random baseline randomly predicts a difficulty level, while the Majority baseline consistently predicts the majority level in the training set.

The experiments are implemented in PyTorch \cite{Ansel_PyTorch_2_Faster_2024} using the HuggingFace~\cite{wolf2020transformers} package, and results are averaged over five independent runs with random seeds.


\section{Results and Discussion}
\label{sec:results_discussion}

\subsection{Balanced DRPS}

Tables \ref{tab:results_racepp} and \ref{tab:results_arc} present the results for the RACE\texttt{++} and ARC datasets, which contain 3 and 7 difficulty levels, respectively.
Balanced DRPS with probabilistic inputs is the main evaluation metric, as these predictions express uncertainty and are particularly valuable in downstream decision-making.
In addition, we report the balanced DRPS computed using degenerate distributions---i.e., where all probability mass is placed entirely on the predicted level.
This setup removes any representation of uncertainty from the predictions, thereby altering the scores for both classification and ordinal regression methods. 
Notably, the baselines and the discretized regression model remain unaffected in this setting, as they do not express uncertainty.
Recall that for both metrics, lower is better.
Furthermore, we include the commonly used but flawed metrics RMSE and accuracy.

\begin{table}[t!]
	\centering
	\caption{\label{tab:results_racepp}Results on RACE\texttt{++} (3 levels)}
	\begin{tabular}{lcccc}\toprule 
		\textbf{Output type} & \textbf{Bal. DRPS $\downarrow$} & \makecell{\textbf{Bal. DRPS $\downarrow$} \\ \textbf{(degenerate)} } & \textbf{RMSE $\downarrow$} &  \textbf{Accuracy $\uparrow$}\\
		\midrule
		Random & 0.893 \gray{$\pm$ 0.009} & 0.893 \gray{$\pm$ 0.009} & 1.024 \gray{$\pm$ 0.005} & 0.334  \gray{$\pm$ 0.003}\\
		Majority & 0.667 \gray{$\pm$ 0.000} & 0.667 \gray{$\pm$ 0.000} & 0.616 \gray{$\pm$ 0.000} & 0.620 \gray{$\pm$ 0.000} \\
		\hdashline\noalign{\vskip 0.5ex}
		Regression & 0.167 \gray{$\pm$ 0.003} & \textbf{0.167} \gray{$\pm$ 0.003} & \textbf{0.391} \gray{$\pm$ 0.004} & 0.853 \gray{$\pm$ 0.003} \\
		Classification & \textbf{0.133} \gray{$\pm$ 0.003} & 0.170 \gray{$\pm$ 0.004} & 0.402 \gray{$\pm$ 0.006} & 0.847 \gray{$\pm$ 0.004} \\
		OR-NN & \textbf{0.131} \gray{$\pm$ 0.004} & \textbf{0.168} \gray{$\pm$ 0.004} & 0.400 \gray{$\pm$ 0.008} & 0.847 \gray{$\pm$ 0.006} \\
		CORAL & 0.201 \gray{$\pm$ 0.003} & 0.185  \gray{$\pm$ 0.004} & 0.476 \gray{$\pm$ 0.018} & 0.782 \gray{$\pm$ 0.017} \\
		CORN & \textbf{0.127} \gray{$\pm$ 0.003} & \textbf{0.164} \gray{$\pm$ 0.001} & 0.397 \gray{$\pm$ 0.004} & 0.851 \gray{$\pm$ 0.003} \\
		\hdashline\noalign{\vskip 0.5ex}
		OrderedLogitNN & \textbf{0.130} \gray{$\pm$ 0.001} & \textbf{0.162} \gray{$\pm$ 0.002} & \textbf{0.384} \gray{$\pm$ 0.005} & \textbf{0.861} \gray{$\pm$ 0.003}\\
		\bottomrule
	\end{tabular}
\end{table}

\begin{table}[t!]
	\centering
	\caption{\label{tab:results_arc}Results on ARC (7 levels)}
	\begin{tabular}{lcccc}\toprule 
		\textbf{Output type} & \textbf{Bal. DRPS $\downarrow$} & \makecell{\textbf{Bal. DRPS $\downarrow$} \\ \textbf{(degenerate)} } & \textbf{RMSE $\downarrow$} &  \textbf{Accuracy $\uparrow$}\\
		\midrule
		Random & 2.296 \gray{$\pm$ 0.017} & 2.296 \gray{$\pm$ 0.017} & 2.730 \gray{$\pm$ 0.006} & 0.144 \gray{$\pm$ 0.002}\\
		Majority & 2.286 \gray{$\pm$ 0.000} & 2.286 \gray{$\pm$ 0.000} & 2.195 \gray{$\pm$ 0.000} & 0.409 \gray{$\pm$ 0.000} \\
		\hdashline\noalign{\vskip 0.5ex}
		Regression & 1.030 \gray{$\pm$ 0.015} & 1.030 \gray{$\pm$ 0.015} & \textbf{1.412} \gray{$\pm$ 0.013} & 0.356 \gray{$\pm$ 0.003} \\
		Classification & 0.720 \gray{$\pm$ 0.003} & 1.038 \gray{$\pm$ 0.005} & 1.522 \gray{$\pm$ 0.013} & \textbf{0.421} \gray{$\pm$ 0.005} \\
		OR-NN & 0.722 \gray{$\pm$ 0.005} & 1.005 \gray{$\pm$ 0.003} & 1.435 \gray{$\pm$ 0.009} & 0.400 \gray{$\pm$ 0.002} \\
		CORAL & 0.963 \gray{$\pm$ 0.003} & 1.440 \gray{$\pm$ 0.020} & 2.083 \gray{$\pm$ 0.033} & 0.156 \gray{$\pm$ 0.005} \\
		CORN & 0.725 \gray{$\pm$ 0.007} & 1.046 \gray{$\pm$ 0.012} & \textbf{1.428} \gray{$\pm$ 0.004} & 0.389 \gray{$\pm$ 0.008} \\
		\hdashline\noalign{\vskip 0.5ex}
		OrderedLogitNN & \textbf{0.674} \gray{$\pm$ 0.004} & \textbf{0.980} \gray{$\pm$ 0.007} & 1.468 \gray{$\pm$ 0.007} & 0.393 \gray{$\pm$ 0.006}\\
		\bottomrule
	\end{tabular}
\end{table}

On RACE\texttt{++} with 3 levels (Table \ref{tab:results_racepp}), the classification model performs comparably to the ordinal methods OR-NN, CORN, and OrderedLogitNN in terms of balanced DRPS.
The discretized regression model, by contrast, shows slightly inferior performance. 
Interestingly, the CORAL model underperforms significantly, likely due to its weight-sharing constraint limiting the NN's capacity. 
Nonetheless, all models substantially outperform the baseline approaches. 
We hypothesize that the small differences in performance across models are due to the limited number of ordinal levels.

The ARC dataset with 7 levels (Table \ref{tab:results_arc}) presents a more challenging setting, as it includes a greater number of difficulty levels and exhibits more pronounced class imbalance. 
Here, the OrderedLogitNN model considerably outperforms all other methods. 
For the remaining methods, the insights are consistent with those observed on RACE\texttt{++}.

When restricting the predictions to degenerate distributions, scores generally deteriorate (see the third column in Tables \ref{tab:results_racepp} and \ref{tab:results_arc}) because balanced DRPS is designed to reward well-calibrated probabilistic predictions while penalizing overconfident, incorrect ones.
This shift brings the regression model’s performance in line with the classification model, OR-NN, and CORN.
OrderedLogitNN again performs on par for RACE\texttt{++} and performs substantially better on ARC.

When considering RMSE and accuracy---metrics that are poorly suited for ordinal prediction tasks (see Section \ref{subsec:metrics})---we observe that models closely related to these objectives unsurprisingly achieve the best performance.
On the RACE\texttt{++} dataset, the results are close and OrderedLogitNN performs comparably to or even slightly better than the regression and classification models.
In contrast, on ARC, the regression approach achieves the lowest RMSE, while classification achieves the highest accuracy.
These results are expected, as the regression method is directly optimized with RMSE loss while the classification method entirely omits the ordinal information, just like the accuracy metric.

\subsection{Confusion matrix}

\begin{figure}[t!]
	\centering
	\begin{subfigure}[b]{0.30\textwidth}
		\centering
		\includegraphics[width=1.0\linewidth]{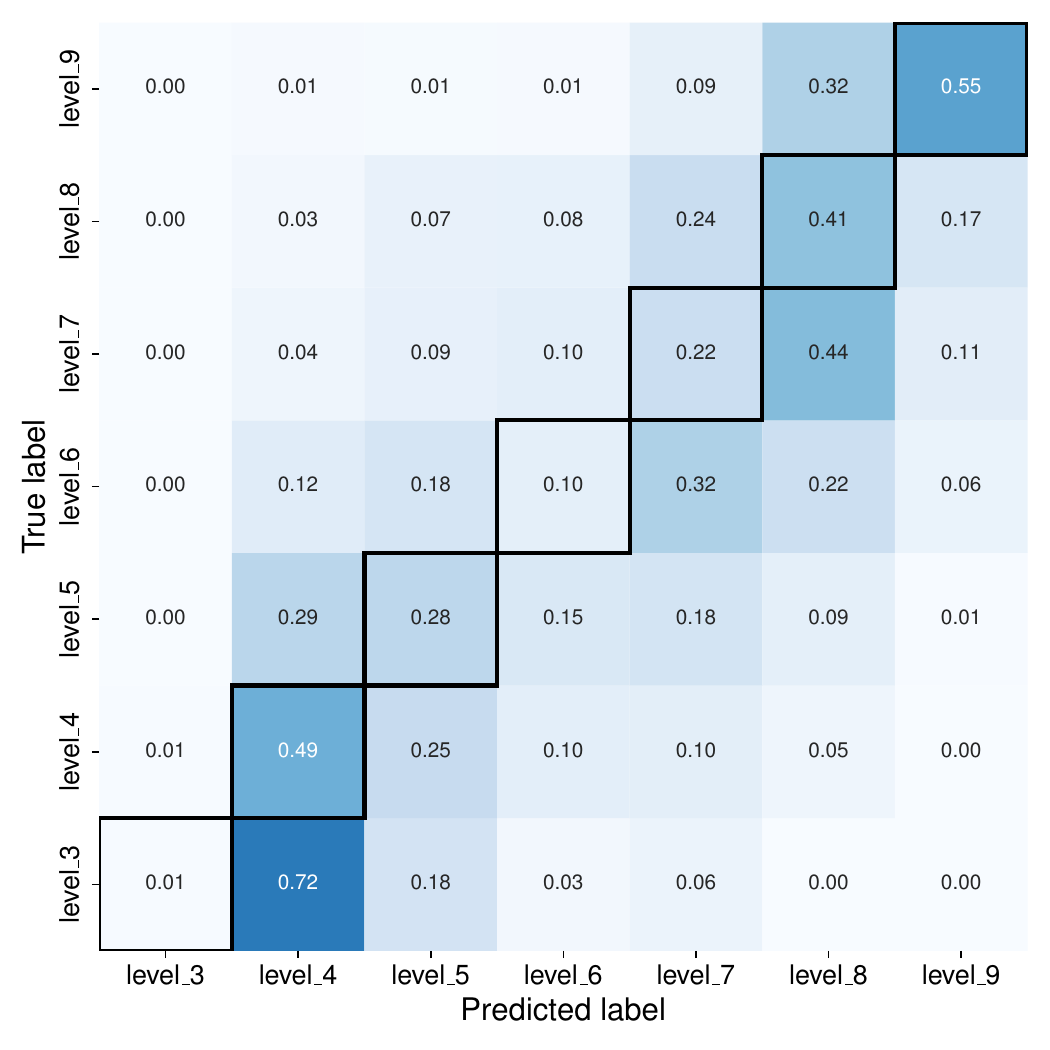}
		\caption{Regression}
		\label{fig:confusion_regression} 
	\end{subfigure}%
\hfill
	\begin{subfigure}[b]{0.30\textwidth}
		\centering
		\includegraphics[width=1.0\linewidth]{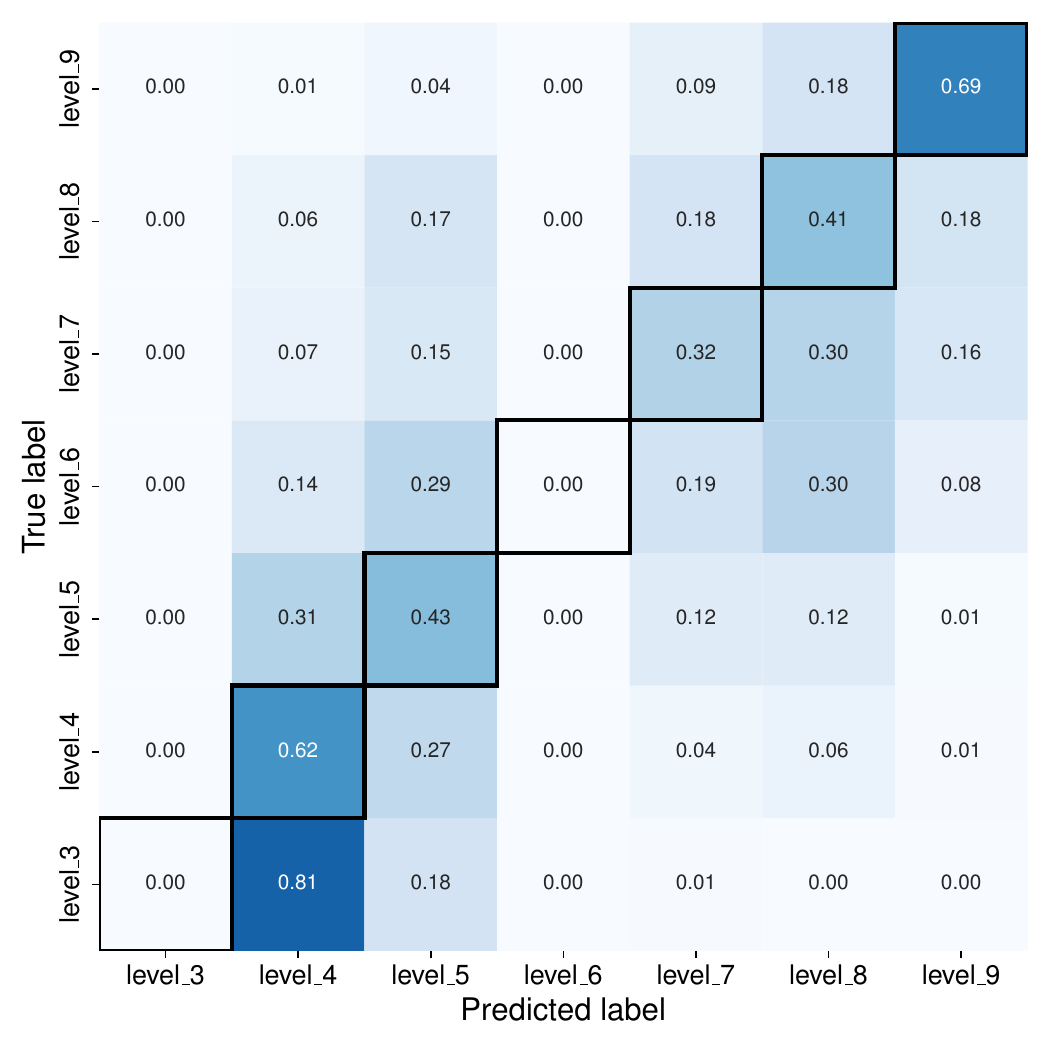}
		\caption{Classification}
		\label{fig:confusion_classification} 
	\end{subfigure}
\hfill
	\begin{subfigure}[b]{0.30\textwidth}
		\centering
		\includegraphics[width=1.0\linewidth]{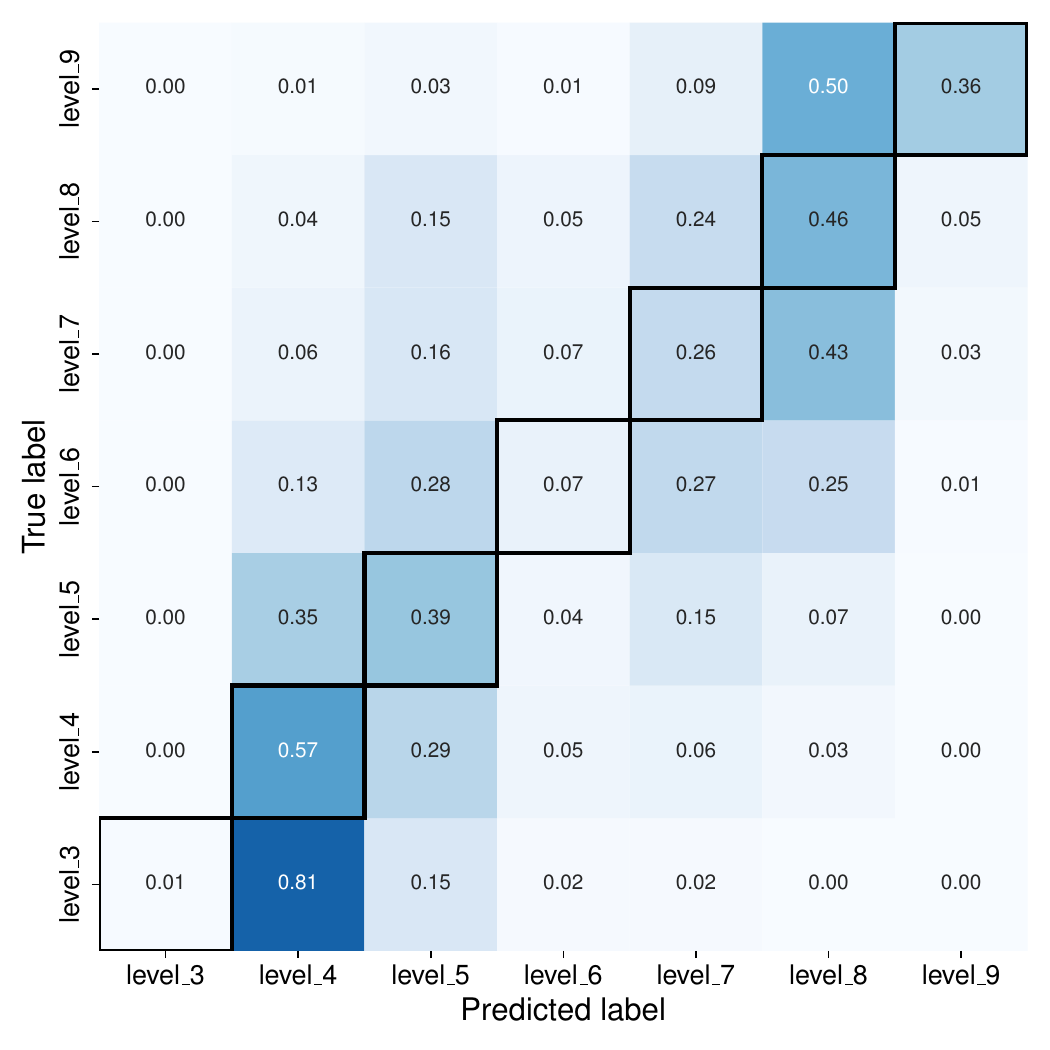}
		\caption{OR-NN}
		\label{fig:confusion_or_nn} 
	\end{subfigure}

	\begin{subfigure}[b]{0.30\textwidth}
		\centering
		\includegraphics[width=1.0\linewidth]{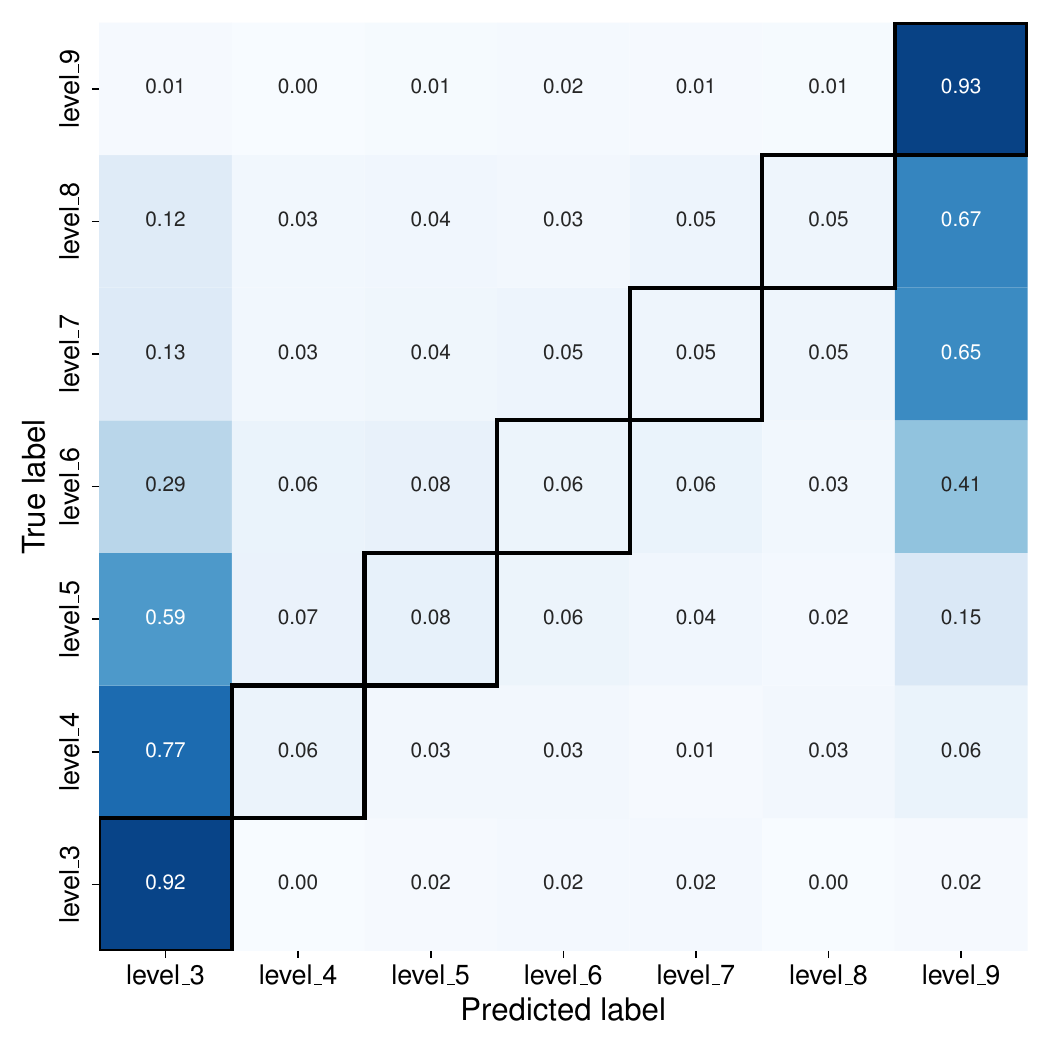}
		\caption{CORAL}
		\label{fig:confusion_coral} 
	\end{subfigure}
\hfill
	\begin{subfigure}[b]{0.30\textwidth}
		\centering
		\includegraphics[width=1.0\linewidth]{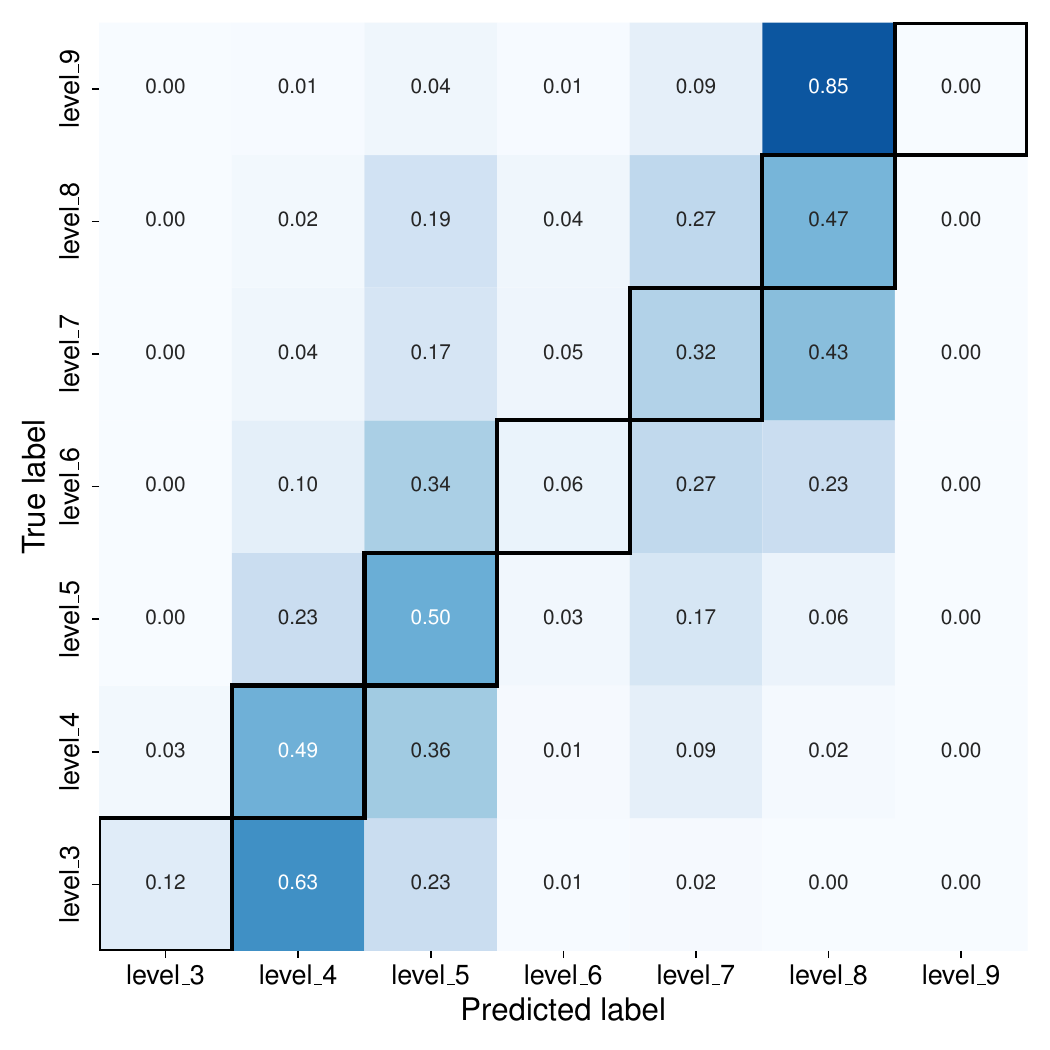}
		\caption{CORN}
		\label{fig:confusion_corn} 
	\end{subfigure}
\hfill
\begin{subfigure}[b]{0.30\textwidth}
	\centering
	\includegraphics[width=1.0\linewidth]{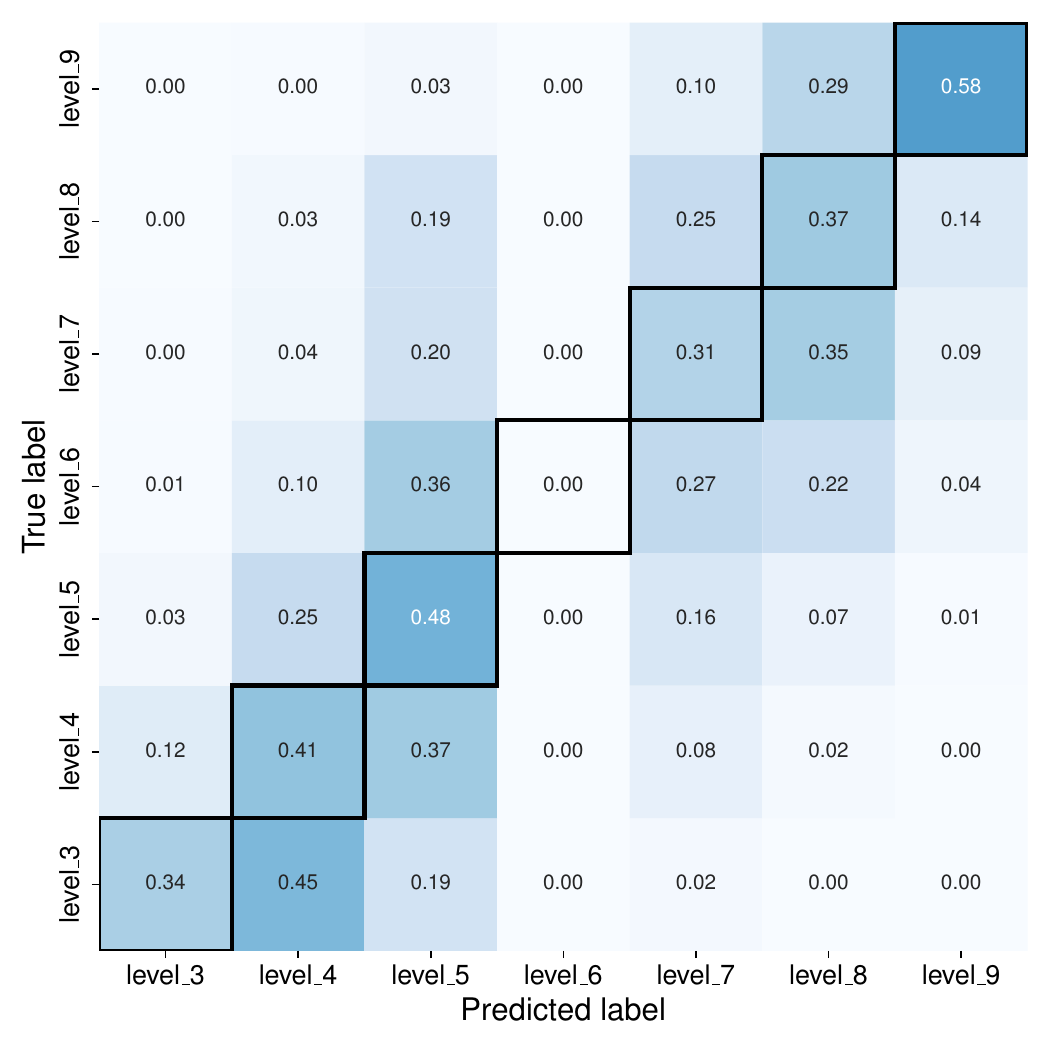}
	\caption{OrderedLogitNN}
	\label{fig:confusion_orderedlogitnn} 
\end{subfigure}
	\caption{Confusion matrices on the ARC dataset for all models, normalized over the true labels (rows).}
	\label{fig:confusion_arc}
\end{figure}

To further investigate the behavior of the models, Figure \ref{fig:confusion_arc} presents confusion matrices for the ARC dataset, the more complex task in this study. 
These matrices are normalized by the true class (i.e., row-wise) and are based on discrete predicted levels---rather than full probability distributions---thus aligning with the evaluation setting of the balanced DRPS with degenerate predictions.

The results indicate the inherent difficulty of the task, as demonstrated by the substantial deviations from the diagonal (highlighted in black).
All models hardly predict level 6 as these observations are least present in the training set.
Notably, the CORAL model exhibits highly atypical behavior, predicting exclusively at the extreme levels (3 and 9).
This pattern suggests that the model fails to converge to a meaningful solution, consistent with its poor performance on the balanced DRPS metric.

Among the remaining models, all except OrderedLogitNN show concentrated errors in specific off-diagonal cells, i.e., predicting level 4 instead of level 3, or level 8 instead of level 9. 
These errors are associated with the outermost classes (levels 3 and 9), which are often neglected entirely by the models. 
In contrast, OrderedLogitNN stands out as the only method that successfully captures both extremes of the ordinal scale, avoiding these consistent misclassifications.


\section{Conclusion}
\label{sec:conclusion}

This study approaches discrete-level QDE through the lens of ordinal regression, reflecting the inherent ordering of difficulty levels from easiest to hardest. 
Prior work in this area has ignored this ordinal structure, both in the choice of modeling paradigms and in the design of evaluation metrics.

To address these gaps, we benchmark three types of model outputs---discretized regression, classification, and ordinal regression---using the balanced DRPS, a novel metric that captures both ordinality and class imbalance. 
Moreover, we propose a novel ordinal regression model OrderedLogitNN, extending the ordered logit model from econometrics to NNs.
We fine-tune the Transformer model BERT on the RACE\texttt{++} and ARC datasets.

Experimental results indicate that OrderedLogitNN considerably outperforms existing methods on more complex tasks while performing comparably on simpler ones, both on probabilistic and degenerate predictions.
Probabilistic predictions express uncertainty and are particularly valuable in downstream human-in-the-loop tasks such as selective prediction or active learning \cite{thuy2023explainability,thuy2024active}.
These findings underscore OrderedLogitNN as a highly attractive approach for discrete-level QDE.
The learning dynamics of OrderedLogitNN deserve further investigation and future work can assess the impact of adapting the weight initializations and learning rate multiplier.

Our work has important practical implications for QDE, as more reliable difficulty estimation enables scalable personalized learning paths in educational platforms.
Furthermore, it has broader relevance for the automated evaluation of assessment content with ordinal labels, such as essay correction.
The balanced DRPS provides a principled foundation for evaluating such systems in production and for future research.
Finally, OrderedLogitNN’s robustness makes it well-suited for integration in educational applications where strong performance over the entire label range is critical.

\begin{acknowledgments}
	We thank the anonymous reviewers for providing valuable feedback.
	This study was supported by the Research Foundation Flanders (FWO) (grant number 1S97022N).
\end{acknowledgments}

\section*{Declaration on Generative AI}

During the preparation of this work, the authors used GPT-4o in order to check grammar and spelling. 
After using this tool, the authors reviewed and edited the content as needed and take full responsibility for the publication’s content. 



\end{document}